\documentclass[letterpaper]{article}
\usepackage[preprint]{aaai2027}
\usepackage[hyphens]{url}
\usepackage{graphicx}
\usepackage{natbib}
\usepackage{caption}
\usepackage{amsmath,amsfonts,amssymb}
\usepackage{booktabs}
\usepackage{multirow}
\usepackage{xspace}
\usepackage{algorithm}
\usepackage{algorithmic}
\usepackage{placeins}

\newcommand{\method}{ReVisIT\xspace}
\newcommand{\bench}{MAAC-Bench\xspace}

\title{Retrieved Images as Visual Thought: \\
Training-Free Multimodal In-Context Learning \\
for the Open-vs-Closed Gap}

\author{
    Bingchen Huang\equalcontrib,
    Zhiling Wang\equalcontrib,
    Yifu Chen\equalcontrib,
    Yuanchao Du
}
\affiliations{
    Meituan\\
    \{huangbingchen, wangzhiling02, chenyifu05, duyuanchao\}@meituan.com
}

\begin{document}
\maketitle

\begin{abstract}
Recent work on Thinking with Images makes vision a dynamic part of reasoning, but does so through generation: the model invokes external tools, synthesizes code, or imagines new imagery, each at the cost of a tool protocol, brittle code, or an expensive training pipeline. A fourth route makes vision dynamic without generating anything, by retrieving labeled exemplar images and reasoning over them, yet it remains underexplored despite being train-free. We present \method, a train-free framework that realizes this retrieval-based route by treating each retrieved image-label pair as a unit of visual thought. \method combines structured class definitions, per-query multimodal retrieval of exemplars, and alternating user/assistant injection of those exemplars before joint multi-attribute decoding, and degrades gracefully to whichever components a task admits. On VL-ICL Bench Fast Open MiniImageNet, Qwen3-VL-30B-A3B with \method reaches \textbf{98.5\%} at 4-shot, statistically indistinguishable from the 72B LLaVA-OneVision SOTA (98.7\%) on this near-saturated task at about 1/2.4 the parameters, while the same backbone without the scaffold sits at chance. The turns layer alone adds 26.1 points to GPT-4.1 on free-form concept induction (Bongard-OpenWorld), and the full stack yields a 4--6 point macro gain across three backbones on \bench, a new license-clean 27-class, 5-attribute benchmark, significant by paired bootstrap on the curator-derived attributes. Component analysis shows that retrieval-plus-turns is the universal lever while structured definitions are need-adaptive, and that 83\% of the retrieval gain comes from retrieval quality rather than from the presence of exemplars. \bench is released with a rubric-grounded LLM verification protocol that replaces author spot-check on subjective attributes.
\end{abstract}

\begin{figure}[!t]
    \centering
    \includegraphics[width=\linewidth]{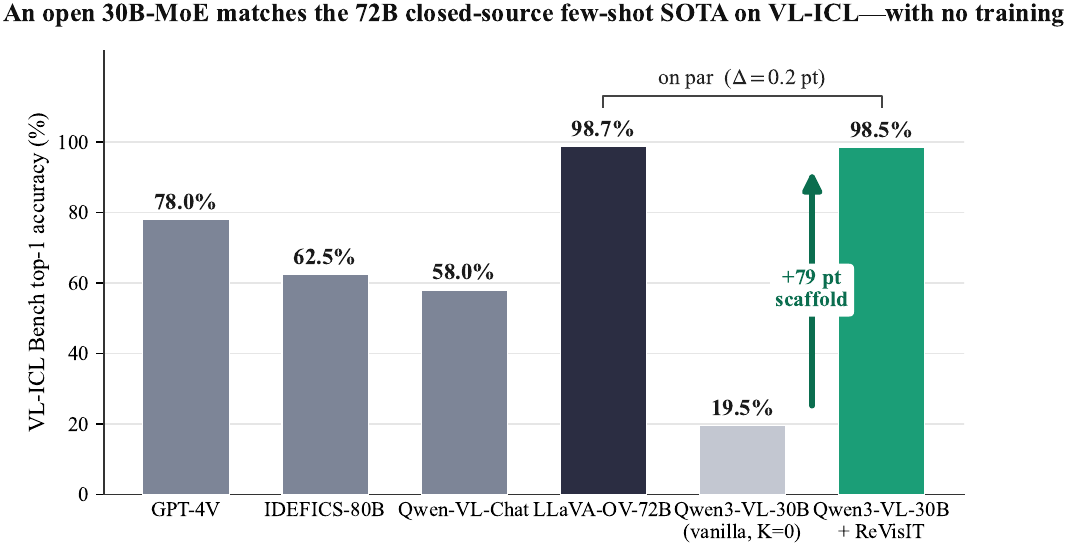}
    \caption{Headline result on the public VL-ICL Bench Fast Open MiniImageNet. An open-source Qwen3-VL-30B-A3B paired with \method reaches \textbf{98.5\%} (4-shot), on par with the 72B LLaVA-OneVision SOTA (98.7\%) at about 1/2.4 the total parameters and 20.5~points above GPT-4V (78.0\%). The same backbone without the \method scaffold sits at chance, isolating the gain to the scaffold rather than the base model.}
    \label{fig:hero}
\end{figure}

\section{Introduction}
\label{sec:intro}

A growing line of work in multimodal AI positions vision not as a static input but as a dynamic, manipulable workspace that a model can act on while it reasons, a paradigm now described as Thinking with Images (TwI)~\cite{zhao2025twi}. This literature has developed along three stages, distinguished by how the model makes vision dynamic. In tool-driven visual exploration, the model orchestrates external visual tools to crop or mark an image (Set-of-Mark~\cite{yang2023som}, ViperGPT~\cite{suris2023vipergpt}). In programmatic visual manipulation, it emits code or sketches that interrogate the image (Visual Sketchpad~\cite{hu2024sketchpad}). In intrinsic visual imagination, it generates internal imagery as a reasoning intermediate (DeepEyes~\cite{deepeyes2025}, Chain-of-Focus~\cite{cof2025}).

These three stages share a limitation as well as a goal. Each makes vision dynamic through generation, and so each carries a cost: a custom tool protocol, unstable code synthesis, or an expensive SFT/RL pipeline that distils the imagination behaviour into the model. This raises a question that the TwI literature has largely left open: is generation actually necessary? What the three stages have in common is not that they generate, but that they reason over visual content absent from the original prompt; tool use crops it, code renders it, and imagination synthesizes it. Retrieval can supply such content too, by pulling labeled exemplar images from a support pool, and it is the one source that synthesizes nothing. Yet this fourth, retrieval-based route to visual dynamism is conspicuously underexplored, even though it is the cheapest and most reproducible of the four and needs no training.

We close this gap with \method (Retrieved Visual In-context Thinking), a train-free framework that realizes the retrieval-based route. \method treats each retrieved image--label pair as a unit of visual thought and places $K$ such pairs as alternating user/assistant turns before the query, so that the model's joint multi-attribute prediction emerges from a chain of $K$ visual demonstrations rather than a chain of generated reasoning steps. The framework has three components: structured class definitions with optional \texttt{confusable\_with} and \texttt{negative\_samples} slots, per-query multimodal retrieval of $K$ exemplars from a labeled support pool, and alternating-turns injection that mirrors how chat-tuned VLMs are trained.

We study \method along three axes. First, we ask whether one retrieval-based design transfers across qualitatively different multimodal tasks: definition-grounded multi-attribute classification, synthetic-name few-shot in-context learning, and free-form concept induction. Second, we ask where its gain comes from, separating the effect of having retrieved exemplars at all from that of retrieving the right ones, and tracing the dependence on shot count and embedder. Third, we ask how far the design narrows the gap between open and closed models, testing whether a moderately sized open VLM with \method can reach the neighborhood of closed frontier systems without any training.

Our contributions are as follows. (1) We introduce \method, a train-free, backbone-agnostic framework with three composable layers (alternating-turns injection, per-query retrieval, and structured definitions) that we position as the retrieval-based member of the Thinking-with-Images design space, the only one needing neither training, tools, nor generated imagery (\S\ref{sec:method}). (2) We show that this one framework transfers across three regimes (\S\ref{sec:experiments}): open Qwen3-VL-30B-A3B with \method reaches \textbf{98.5\%} on VL-ICL, on par with the 72B LLaVA-OneVision SOTA at about 1/2.4 the parameters, adds 26.1 points on Bongard-OpenWorld, and yields a 4--6 point macro gain across three backbones on \bench. (3) We analyze the layers, identifying retrieval-plus-turns as the dominant, backbone-universal lever and attributing 83\% of the retrieval gain to retrieval quality rather than the mere presence of exemplars (\S\ref{sec:exp-buildup}). (4) We release \bench, a license-clean 27-class, 5-attribute art-classification benchmark with a rubric-grounded LLM verification protocol that replaces author spot-check by judging ensemble labels against externally sourced definitions (\S\ref{sec:bench}).

\section{Related Work}
\label{sec:related}

\paragraph{Thinking with Images.}
\citet{zhao2025twi} survey work that treats vision as a dynamic workspace and organize it into three stages distinguished by how the model makes vision dynamic: tool-driven exploration, in which the model orchestrates external vision tools to crop, mark, or zoom (Set-of-Mark~\cite{yang2023som}); programmatic manipulation, in which it emits code or sketches that interrogate the image (ViperGPT~\cite{suris2023vipergpt}, Visual Sketchpad~\cite{hu2024sketchpad}); and intrinsic visual imagination, in which it generates internal imagery as reasoning intermediates, typically via reinforcement learning (DeepEyes~\cite{deepeyes2025}, Chain-of-Focus~\cite{cof2025}). All three produce visual dynamism by generation, and the more capable stages require expensive SFT/RL pipelines. \method occupies the missing retrieval-based position in this space: it makes vision dynamic by retrieving labeled exemplars and arranging them as alternating chat turns, the only route in the family that is train-free. We do not claim a new mechanism, since retrieval-augmented in-context learning is established (below); our aim is to characterize and benchmark retrieval as the underexplored, train-free member of the family against the generation-based stages.

\paragraph{In-context and retrieval-augmented learning.}
A line of work establishes that frontier VLMs can perform in-context few-shot classification (Flamingo~\cite{alayrac2022flamingo}, OpenFlamingo~\cite{awadalla2023openflamingo}, IDEFICS~\cite{laurencon2024idefics}, LLaVA-OneVision~\cite{li2024llava_onevision}, VL-ICL Bench~\cite{zong2024vlicl}), and contrastive models cast classification as image-to-text similarity (CLIP~\cite{radford2021clip}, OpenCLIP~\cite{ilharco2021openclip}, ALIGN~\cite{jia2021align}, CoCa~\cite{yu2022coca}), though the latter degrade when a short label name stands in for a paragraph of meaning. Retrieving query-relevant exemplars to populate the prompt is also well established: semantically retrieved examples beat random selection in NLP~\cite{liu2022goodexamples}, RICES-style retrieval selects support images by frozen-encoder similarity~\cite{alayrac2022flamingo}, and Re-ViLM~\cite{yang2023revilm} and MRAG-Bench~\cite{hu2024mragbench} study retrieval-augmented captioning and VQA. We build on this line rather than claim retrieval-for-ICL as new. Our additions are to combine per-query retrieval with first-class structured definitions and joint multi-attribute decoding, which prior retrieval-ICL work does not, and to inject exemplars as alternating user/assistant turns rather than a flat support paragraph (\S\ref{sec:method}), a format we verify matches flat concatenation rather than treating it as load-bearing (\S\ref{sec:exp-ablation}). Where classical RAG for medical-image~\cite{ramesh2023med} and document~\cite{lewis2020rag} classification embeds retrieved evidence as text, we weave multimodal image--label pairs into the chat-turn stream. A separate line freezes the VLM and tunes lightweight components for contrastive encoders, via learned prompts (CoOp~\cite{zhou2022coop}, CoCoOp~\cite{zhou2022cocoop}), a cached index (Tip-Adapter~\cite{zhang2022tipadapter}), or prompt-drift regularizers (KgCoOp~\cite{yao2023kgcoop}, PromptSRC~\cite{khattak2023promptsrc}); \method instead tunes nothing and targets generative chat-tuned VLMs whose in-context behaviour is the lever, grounding classification on textual definitions rather than learned tokens.

\paragraph{Art classification and benchmarks.}
WikiArt-based classification~\cite{saleh2016wikiart, strezoski2018omniart} and aesthetic-rating datasets (AVA~\cite{murray2012ava}, AADB~\cite{kong2016aadb}) treat single-attribute prediction, while ELEVATER~\cite{liu2024elevater} and Bongard-OpenWorld~\cite{wu2024bongardow} test single-attribute few-shot classification and concept induction. \bench packages Movement, Genre, Medium, Composition, and Mood as a joint multi-attribute few-shot benchmark with copyright-clean provenance.

\section{Method and Benchmark}
\label{sec:method}

\method, short for Retrieved Visual In-context Thinking, makes vision dynamic by retrieval: it treats each retrieved image--label pair as a unit of visual thought and reasons over a chain of such pairs rather than over generated intermediates. This section formalizes the framework and pipeline, then introduces the \bench benchmark.

\subsection{Problem and pipeline}

\paragraph{Problem.}
Given a taxonomy $\mathcal{T} = \{(c_i, n_{ij}, d_{ij})\}$ of categories $c_i$, class names $n_{ij}$, and text definitions $d_{ij}$; a support pool $\mathcal{S} = \{(x_s, y_s)\}$ of labeled positives; and a query image $x_q$, predict per-category labels $\hat{y}_q$ subject to per-category cardinality limits. No gradient updates on $\mathcal{T}\cup\mathcal{S}$ are performed.

\paragraph{Pipeline (Fig.~\ref{fig:pipeline}; pseudocode in Appendix~\ref{sec:supp-algo}).}
\method has five stages. Indexing encodes every $(x_s, y_s)\in\mathcal{S}$ once with a multimodal embedder $\phi$ (default Qwen3-VL-Embedding-8B) into $e_s=\phi(x_s,y_s)\in\mathbb{R}^D$. Per-query retrieval embeds $e_q=\phi(x_q,\emptyset)$ and selects the top-$K$ exemplars by cosine similarity. Prompt compilation assembles a structured three-block system prompt, comprising \texttt{<instructions>}, \texttt{<description>} (each $(c_i,n_{ij},d_{ij})$ on one line, with optional \texttt{confusable\_with} and \texttt{negative\_samples} clauses rendered if present, else the free-text definition as-is), and \texttt{<category>} (the label space), with the $K$ exemplars inlined as alternating user/assistant turn pairs. Generation greedy-decodes the query JSON \texttt{\{"tags":\{$c_i$:[labels]\}\}}. Finally, confidence and validation assign each predicted label $\ell$ a score $\mathrm{conf}(\ell)=\mathrm{softmax}(\mathrm{logits}_t)[\mathrm{argmax}]$ at the first token containing $\ell$, then filter to the per-category whitelist and truncate to $\mathrm{limit}(c_i)$ labels.

\begin{figure}[t]
    \centering
    \includegraphics[width=\linewidth]{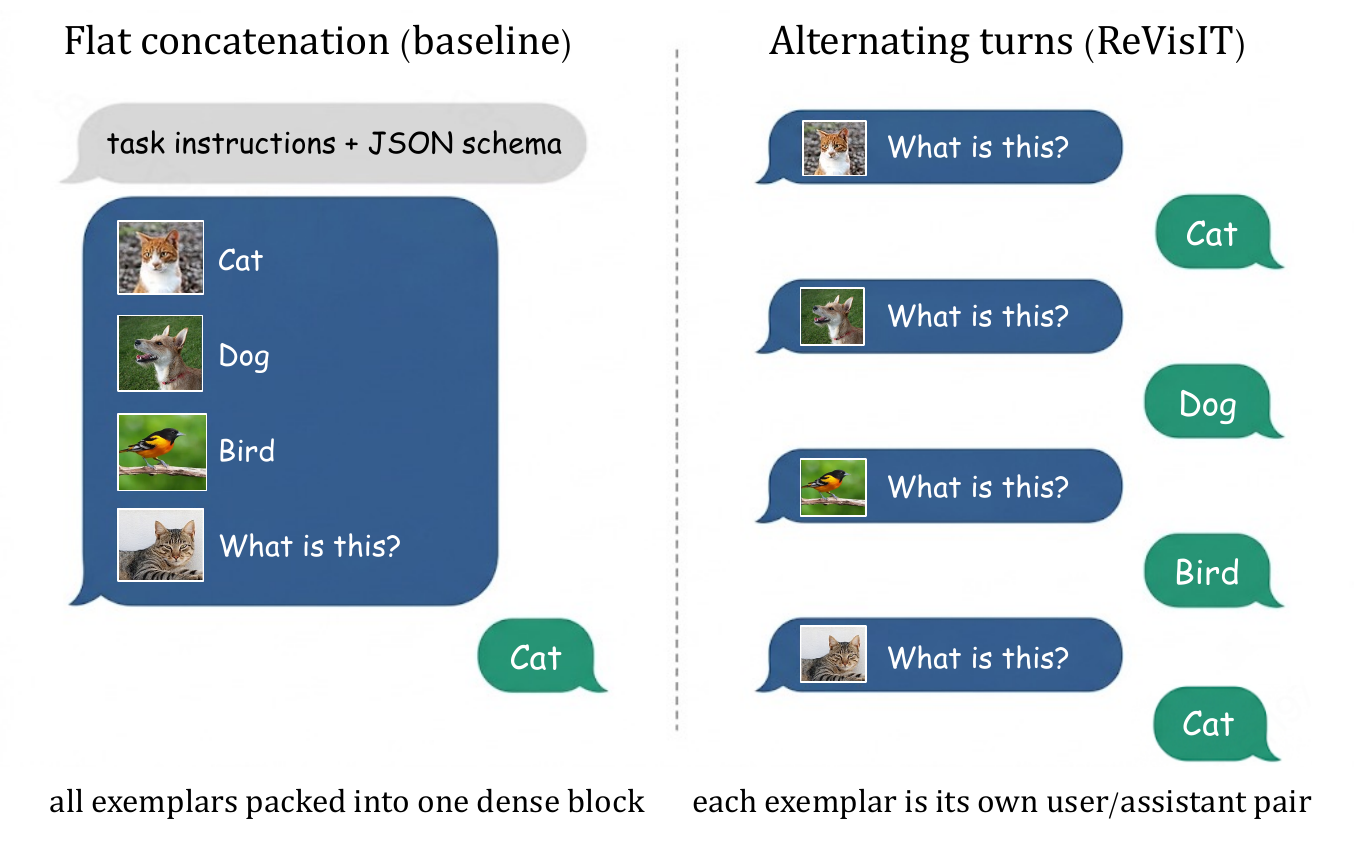}
    \caption{Exemplar injection formats. \textbf{Left}: standard flat concatenation packs all $K$ exemplars into a single support turn, so the model resolves the query against one dense block. \textbf{Right}: \method's alternating turns inline each exemplar as its own user/assistant pair, mirroring how chat-tuned VLMs are trained.}
    \label{fig:altturns}
\end{figure}

\paragraph{Alternating-turns exemplar injection.}
A standard practice is to concatenate $K$ exemplars into a single support paragraph (Fig.~\ref{fig:altturns}, left). We instead inline each exemplar as a separate user/assistant turn pair, where the user message carries the task prompt plus the exemplar image and the assistant message carries the exemplar's gold JSON answer (Fig.~\ref{fig:altturns}, right), and append the query as one more user turn. This mirrors how multi-turn chat-tuned VLMs~\cite{alayrac2022flamingo, laurencon2024idefics, wang2024qwen2vl} are trained and composes naturally with the assistant-role answer. We adopt it for this alignment rather than for accuracy: a turns-versus-flat ablation (\S\ref{sec:exp-ablation}) finds the two formats statistically indistinguishable on a strong backbone, both at 100\% JSON adherence, so the turn structure is a deliberate, low-cost design choice rather than a load-bearing trick. Whether it helps weaker open backbones, where flat multi-image context more often degrades, is left to future work.

\begin{figure*}[!t]
    \centering
    \includegraphics[width=0.92\linewidth]{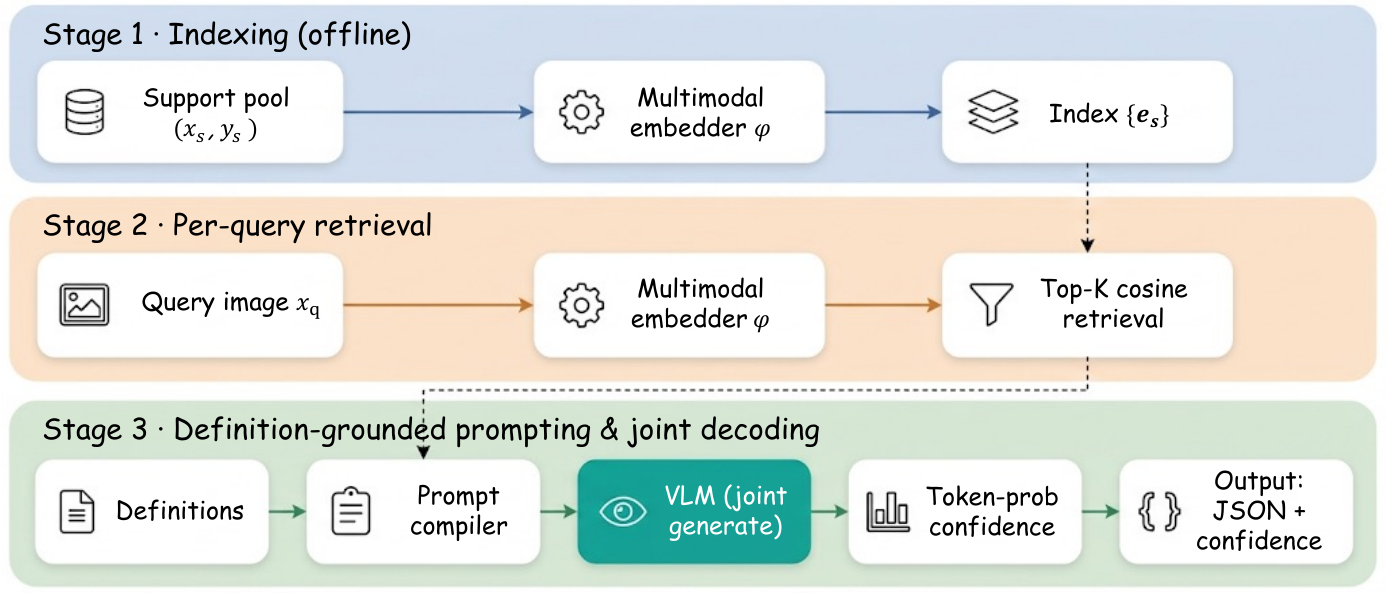}
    \caption{The \method pipeline. Stage~1 indexes the support pool once with a multimodal embedder $\phi$. Stage~2 embeds the query and retrieves top-$K$ exemplars by cosine similarity. Stage~3 compiles a structured three-block prompt with alternating user/assistant exemplar turns, runs a single joint VLM generation, and assigns per-label confidences from token-level softmax probabilities. Only the VLM step is non-deterministic; everything else is a deterministic function of the inputs.}
    \label{fig:pipeline}
\end{figure*}

\subsection{The \bench benchmark}
\label{sec:bench}

\paragraph{Why a new benchmark.}
Existing few-shot benchmarks test single-attribute classification (ELEVATER~\cite{liu2024elevater}, VL-ICL~\cite{zong2024vlicl}) or concept induction (Bongard-OpenWorld~\cite{wu2024bongardow}); none tests joint multi-attribute prediction with definition-grounded supervision, the regime where \method's design choices matter most. \bench fills this gap with small, copyright-clean, curator-rich art-historical data.

\paragraph{Sources and licensing.}
\bench draws from three public sources: Met Museum Open Access (CC0; pre-1928 backbone), National Gallery of Art Open Access (CC0; Abstract Expressionism/Color-Field, CC0 since 2012), and Wikimedia Commons (PD-in-source-country; early-20c European avant-garde, artist died ${\geq}70$ years ago). Movement is derived from a curated artist-to-movement mapping; derived labels are released CC-BY 4.0 with links to the original sources.

\paragraph{Attributes (27 classes total).}
\textbf{Movement} (12): Renaissance, Baroque, Romanticism, Impressionism, Post-Impressionism, Pointillism, Fauvism, Futurism, Suprematism, Surrealism, Abstract Expressionism, Vorticism. \textbf{Genre} (6: Portrait, Landscape, Still Life, \dots), \textbf{Medium} (5: Oil, Watercolor, Fresco, Drawing, Print), \textbf{Composition} (4: Centered, Triangular, Diagonal, Asymmetric), and \textbf{Mood} (5: Serene, Dramatic, Melancholic, Joyful, Mysterious) complete the 27 classes. Movement, Genre, and Medium are derived from curator fields where available; Composition and Mood are annotated by a three-model VLM ensemble (GPT-4.1, GLM-4V-Plus, Doubao-1.5-vision-pro-32k) with majority vote ($\geq 2/3$) as gold; items without majority are excluded. Pairwise Cohen's $\kappa$ is only moderate (best-pair upper bound: Composition $\kappa_{\max}{=}0.43$, Mood $\kappa_{\max}{=}0.54$; the mean is lower, with the full heatmap in Appendix~\ref{sec:supp-kappa}), so we treat Composition and Mood as diagnostic throughout. Ensemble-gold majority rates are 85.6\% and 81.7\%.

\paragraph{Rubric-grounded LLM verification.}
\label{sec:bench-verify}
Because none of this paper's authors is a domain expert in art-historical composition or affect, author spot-check would inject the wrong calibration. We instead run a rubric-grounded LLM verifier as an independent quality gate: the verifier (GPT-4.1, run separately from its annotator role) receives the image, the ensemble's proposed label, the externally sourced authoritative definition (Tate glossary, \emph{Encyclopedia Britannica}, or Wikipedia art-history pages), and the alternative classes' definitions, and returns a yes/no decision plus a proposed alternative on rejection. Of the Composition labels, 89.3\% (117/131) survive verification; for Mood the rate is 63.2\% (79/125). The lower Mood rate is itself diagnostic, as 28 of 46 Mood rejections produce no clear alternative, which the verifier attributes to boundary fuzziness among Serene, Melancholic, and Mysterious, a known difficulty of affect labeling that we report rather than smooth over. To our knowledge this is the first systematic alternative to author spot-check on an LLM-annotated visual benchmark.

\paragraph{Splits.}
We target a uniform $8/5$ query/support split per movement (Surrealism is $8/2$, constrained by public-domain availability). Total: 96 query / 57 support images; the robustness analysis (\S\ref{sec:findings}) expands the curator-derived attributes to 236 queries (Appendix~\ref{sec:supp-stats}).

\section{Experiments}
\label{sec:experiments}

\subsection{Setup}

We test whether \method's layered design is a portable upgrade along three axes that activate different subsets of its layers: synthetic-name fast-binding (VL-ICL, \S\ref{sec:exp-vlicl}; retrieval and turns), definition-grounded multi-attribute classification (\bench, \S\ref{sec:exp-maac}; full stack), and label-free concept induction (Bongard-OpenWorld, \S\ref{sec:exp-bongard}; turns alone). What transfers is the compositional design rather than a fixed prompt.

We evaluate on three benchmarks. VL-ICL Bench Fast Open MiniImageNet~\cite{zong2024vlicl} has 200 queries and 5-way open-ended classification with synthetic concept names (blicket, dax, and so on) fast-binding to ImageNet categories, each query carrying its own 25-image support pool; we evaluate $K\in\{0,1,2,4\}$ shots per class, where the synthetic names remove any pretraining-vocabulary advantage so the $K{=}0$ row is the scaffold-free base model. \bench (\S\ref{sec:bench}) has 96 query and 57 support images over 5 attributes and 27 classes, scored as top-1 accuracy on majority-vote gold and macro-averaged over attributes. Bongard-OpenWorld is free-form concept induction with no symbolic labels.

Three VLMs span the closed and open axis: GPT-4.1, Doubao-1.5-vision-pro-32k, and the open-source Qwen3-VL-30B-A3B-Instruct~\cite{wang2024qwen2vl} (MoE, about 3B active, single A100~80GB). GLM-4V-Plus serves only as an ensemble annotator for \bench and not as an evaluated backbone, because in pilot runs it failed to format multi-image in-context exemplars reliably and degraded under the scaffold; we exclude it from the backbone comparison to avoid conflating a backbone limitation with the framework. Each backbone runs in vanilla mode (structured prompt only, $K{=}0$) and \method mode (definitions, Qwen3-VL-Embedding-8B cosine retrieval, and $K$-shot alternating-turns injection), with greedy decoding throughout; vanilla uses the identical three-block prompt minus exemplars, isolating the retrieval contribution. The full stack runs on \bench, VL-ICL uses retrieval and turns (its 25-image per-query pool makes retrieval near-trivial), and Bongard uses the turns layer alone.

\subsection{Public benchmark: VL-ICL Bench}
\label{sec:exp-vlicl}

\begin{table}[!tb]
\caption{VL-ICL Bench Fast Open MiniImageNet top-1 accuracy across $K\in\{0,1,2,4\}$ shots per class. \emph{Ours} (rows 5--7) are \method's joint output. Rows 1--4 are published numbers from~\cite{zong2024vlicl, li2024llava_onevision} for context. The $K{=}0$ column is the scaffold-free base model (no exemplars): with synthetic names, the open Qwen3-VL-30B alone sits at chance ($0.195$, 5-way), so the $K{=}4$ result is attributable to \method's scaffold rather than the base model's prior knowledge. \textbf{Bold} marks best in column; \underline{underline} second-best.}
\centering
\footnotesize
\setlength{\tabcolsep}{4pt}
\resizebox{\columnwidth}{!}{%
\begin{tabular}{lcccc}
\toprule
Method & $K{=}0$ & $K{=}1$ & $K{=}2$ & $K{=}4$ \\
\midrule
\multicolumn{5}{l}{\emph{Published~\cite{zong2024vlicl}}} \\
GPT-4V                                & --    & --    & --    & 0.780 \\
LLaVA-OneVision-72B~\cite{li2024llava_onevision} & --    & --    & --    & 0.987 \\
IDEFICS-80B~\cite{laurencon2024idefics}          & --    & --    & --    & 0.625 \\
Qwen-VL-Chat                          & --    & --    & --    & 0.580 \\
\midrule
\multicolumn{5}{l}{\emph{Ours} (\method)} \\
\method (GPT-4.1)              & --    & \underline{0.965} & \textbf{1.000} & \textbf{1.000} \\
\method (Doubao-1.5)           & --    & \textbf{0.975}    & \underline{0.995} & \textbf{1.000} \\
\method (Qwen3-VL-30B, open)   & 0.195 & 0.800             & 0.950             & \underline{0.985} \\
\bottomrule
\end{tabular}}
\label{tab:vlicl}
\end{table}

VL-ICL's synthetic class names are absent from any VLM's pretraining, which makes it the cleanest available probe of in-context label binding, though it controls for lexical rather than visual familiarity since MiniImageNet derives from ImageNet. Table~\ref{tab:vlicl} reports our results alongside published baselines.

The headline result is that an open 30B-MoE matches the 72B SOTA. At $K{=}4$, Qwen3-VL-30B-A3B (about 3B active of 30B total) with \method reaches \textbf{0.985}, statistically indistinguishable from the 72B LLaVA-OneVision SOTA (0.987) at about 1/2.4 the total parameters on this near-saturated task: the 0.2-point gap is well under one query of the 200-query set, so we read it as a tie rather than a win. The same backbone scores only 0.195 (chance) at $K{=}0$, and since the synthetic names guarantee no prior knowledge of the concepts, the resulting 79-point lift is attributable to the scaffold rather than the base model. Because each query's support pool is only 25 images, this result mainly exercises the alternating-turns layer; retrieval quality, the framework's main lever, is stressed instead on \bench below. The frontier backbones saturate quickly (GPT-4.1 at 1.000 for $K{\geq}2$, Doubao-1.5 at 1.000 for $K{=}4$), whereas the open Qwen3-VL gains monotonically ($0.195\to0.800\to0.950\to0.985$), so for smaller open VLMs each additional retrieved exemplar remains a high-value lever.

\subsection{\bench per-attribute results}
\label{sec:exp-maac}

\begin{table*}[!tb]
\caption{\bench top-1 accuracy across 5 attributes. \method (rows 4--6) adds Qwen3-VL-Embedding-8B cosine retrieval + 5-shot in-context exemplars on top of the same vanilla VLM (rows 1--3). Row 7 isolates the retrieval-quality contribution by replacing cosine retrieval with random sampling. \textbf{Bold} marks best in column.}
\centering
\small
\setlength{\tabcolsep}{6pt}
\begin{tabular}{lcccccc}
\toprule
Method & Macro & Movement & Genre & Medium & Composition & Mood \\
\midrule
GPT-4.1 (vanilla)                                  & 0.721 & 0.573 & 0.646 & \textbf{0.844} & 0.776 & 0.766 \\
Doubao-1.5-vision-pro-32k (vanilla)                & 0.652 & 0.406 & 0.604 & 0.792 & 0.718 & 0.740 \\
Qwen3-VL-30B-A3B (open, vanilla)                   & 0.635 & 0.427 & 0.615 & 0.750 & 0.694 & 0.688 \\
\midrule
\textbf{\method (GPT-4.1)}                         & \textbf{0.762} & \textbf{0.698} & 0.677 & 0.833 & 0.812 & \textbf{0.792} \\
\method (Doubao-1.5)                               & 0.706 & 0.479 & \textbf{0.750} & 0.823 & 0.765 & 0.714 \\
\method (Qwen3-VL-30B, open)                       & 0.689 & 0.500 & 0.635 & \textbf{0.844} & 0.765 & 0.703 \\
\midrule
\method (GPT-4.1, \emph{random retrieval})         & 0.728 & 0.583 & 0.646 & 0.833 & \textbf{0.835} & 0.740 \\
\bottomrule
\end{tabular}
\label{tab:headline}
\end{table*}

\subsection{Component build-up and ablations}
\label{sec:exp-buildup}

A multi-component method invites the objection that it is three unrelated tricks under one name, so we add the layers one at a time on \bench, on both GPT-4.1 and the open Qwen3-VL-30B, holding everything else fixed (full table in Appendix~\ref{sec:supp-buildup}). Starting from a bare label-space-only configuration (GPT-4.1 0.732, Qwen3-VL-30B 0.610), two readings follow. The exemplar layer of retrieval and alternating turns is the load-bearing, backbone-universal component, adding 4.1 points on GPT-4.1 (to 0.762) and 5.4 points on Qwen3-VL-30B (to 0.689) over the definitions-only configuration. Structured definitions, by contrast, are need-adaptive: they help the open model that lacks domain priors (+2.5) but are slightly counterproductive for the frontier model that already knows the taxonomy ($-$1.1), a coherent pattern rather than an arbitrary stack. The definitions layer also provides interpretability and zero-shot deployability to any new taxonomy without retraining, independent of its accuracy effect.

We then decompose the load-bearing layer.\label{sec:exp-ablation} To separate which exemplars are present from whether exemplars are present at all, we replace cosine top-$K$ retrieval with uniform random sampling, holding the model, structured prompt, and turns format constant (Table~\ref{tab:headline}, row 7). On the GPT-4.1 macro, vanilla scores 0.721, \method with random 5-shot scores 0.728 (+0.7, about 17\% of the gain), and \method with cosine 5-shot scores 0.762 (+4.1, the remaining 83\%); the retrieval embedder, not the few-shot slot, is the load-bearing component. We further isolate the turns format by re-running GPT-4.1 with all five exemplars concatenated into a single flat user message (RICES-style), holding the retrieved content fixed. The two formats are statistically indistinguishable on the 236-query set (flat 0.796 vs.\ turns 0.800 macro; on curator-derived attributes flat is marginally higher at 0.816 vs.\ 0.799), both at 100\% clean-JSON adherence. We therefore do not claim the turn structure as load-bearing; on a strong backbone the retrieval layer carries the gain, and we adopt alternating turns because it mirrors the chat-tuned training format, leaving its effect on weaker open backbones to future work.

Finally, we check that the gain comes from the VLM reasoning over the retrieved exemplars rather than from the retrieval embedding alone, by running two parameter-free probes that use only the Qwen3-VL-Embedding similarities with no VLM call (Table~\ref{tab:probe}). Both fall well short of the full scaffold: nearest-prototype reaches 0.48 and kNN-vote 0.562 on the curator-derived attributes, versus 0.799 for \method on the same retrieved exemplars. The similarities locate relevant exemplars but cannot themselves make the joint multi-attribute decision, so it is the in-context reasoning, not the nearest-neighbour vote, that produces the gain.

\begin{table}[tb]
\caption{Embedding-only probes vs.\ \method on the 236-query curator-derived attributes (macro over Movement, Genre, Medium; same Qwen3-VL-Embedding-8B). The probes use the embedding similarities directly with no VLM call; \method runs the full GPT-4.1 scaffold over the same retrieved exemplars.}
\centering
\small
\setlength{\tabcolsep}{6pt}
\begin{tabular}{lc}
\toprule
Method (embedder = Qwen3-VL-Embedding-8B) & Macro (3 attr) \\
\midrule
Nearest-prototype (per-class mean support) & 0.48 \\
kNN-vote (top-5 support, majority)         & 0.562 \\
\midrule
\textbf{\method (GPT-4.1, full scaffold)}  & \textbf{0.799} \\
\bottomrule
\end{tabular}
\label{tab:probe}
\end{table}

\subsection{Analysis}
\label{sec:findings}

The \bench gain is stable across architecturally diverse backbones, at 4.1 points on GPT-4.1, 5.4 on Doubao-1.5, and 5.4 on the open Qwen3-VL-30B on the 96-query core set. To address the modest benchmark size, we expanded the three curator-derived attributes (Movement, Genre, Medium) to 236 queries by adding 140 license-clean images from the Metropolitan Museum and the National Gallery of Art (Appendix~\ref{sec:supp-stats}) and recomputed a paired bootstrap with 10,000 resamples. All three backbones stay significant with tighter intervals on the larger set: GPT-4.1 $+3.7\,[+1.7,+5.8]$, Doubao-1.5 $+6.1\,[+3.5,+8.8]$, and Qwen3-VL-30B $+5.5\,[+3.0,+8.2]$ points, all with $P(\text{gain}>0)>0.99$. We anchor significance on these curator-derived attributes and leave Composition and Mood at the 96 diagnostic queries, where there is no expansion gold and agreement is lower. Table~\ref{tab:provenance} reports accuracy split by label provenance, so that the curator-derived gains, on which we base our claims, are not blended with the lower-agreement model-derived attributes. The same portability recurs on VL-ICL, where every backbone with \method reaches at least 0.985 at $K{=}4$, so the pattern is a portable upgrade rather than a backbone-specific trick.

\begin{table}[tb]
\caption{\bench macro accuracy split by label provenance: curator-derived (Movement, Genre, Medium; 236 queries) vs.\ model-derived (Composition, Mood; 96 queries). \method's gains are significant only on the curator-derived attributes (paired bootstrap); the model-derived columns are reported for completeness and read as diagnostic.}
\centering
\small
\setlength{\tabcolsep}{4pt}
\begin{tabular}{lcccc}
\toprule
 & \multicolumn{2}{c}{Curator (Mv/Ge/Md)} & \multicolumn{2}{c}{Model (Co/Mo)} \\
Backbone & vanilla & \method & vanilla & \method \\
\midrule
GPT-4.1      & 0.763 & \textbf{0.799} & 0.771 & 0.802 \\
Doubao-1.5   & 0.653 & \textbf{0.713} & 0.729 & 0.740 \\
Qwen3-VL-30B & 0.651 & \textbf{0.706} & 0.722 & 0.734 \\
\bottomrule
\end{tabular}
\label{tab:provenance}
\end{table}

Most of this gain comes from retrieval quality. The random-versus-cosine ablation decomposes the 4.1-point GPT-4.1 macro gain into 0.7 points (about 17\%) from the presence of exemplars and 3.4 points (about 83\%) from retrieving the right ones. The decomposition is specific to \bench: on VL-ICL each query carries its own 25-image pool, making retrieval near-trivial, so that benchmark exercises the turns layer and offers complementary rather than redundant evidence. The effect is also attribute-dependent, carried by Movement and Mood, while on Composition and Medium random retrieval matches or slightly exceeds cosine, consistent with those attributes relying on surface cues that any exemplar supplies. Where category boundaries are subtle, then, a stronger embedder is a more cost-effective lever than more exemplars. On the open-versus-closed question, the open Qwen3-VL with \method (0.689 macro on \bench, 0.985 on VL-ICL) surpasses vanilla Doubao-1.5 (0.652) and ties the 72B SOTA on VL-ICL, making a 30B open model with \method a credible and far cheaper substitute for mid-tier frontier VLMs.

Two further checks bound the result. Because \bench is fine art, a query and an exemplar can share an artist, which could inflate a retrieval method (31\% of queries retrieve a same-artist support in their top-5). On the 163 same-artist-free queries within the 236-query expanded set, the open Qwen3-VL-30B gain is preserved (+5.5 points, matching its bootstrap estimate), while GPT-4.1's shrinks to +1.6 points (from a 0.781 subset vanilla, above its 0.763 full-set vanilla because the subset drops the hardest same-artist confusions); much of GPT-4.1's gain is therefore same-artist retrieval, whereas the open backbone's is leakage-robust, and exact-duplicate leakage is zero. At $K{=}5$, \method incurs a $2.1\times$ token and $1.7\times$ latency overhead over vanilla on GPT-4.1. Across all seven methods, Medium is the easiest attribute (0.69--0.84) and Movement the hardest (0.41--0.70), reflecting the difficulty of distinguishing visually similar twentieth-century avant-garde movements such as Suprematism and Vorticism, and this stable cross-VLM ordering is \bench's main diagnostic value. Movement errors follow an anchor-to-canonical bias dominated by Abstract Expressionism misread as Post-Impressionism (14 of 25 Rothko- and Newman-style queries), a genuinely hard class on which even vanilla GPT-4.1 scores only 0.04. \method still lifts the other sparse movements substantially (Pointillism +44, Fauvism +27, Romanticism, Futurism, and Suprematism about +25 each; Table~\ref{tab:permovement}), so the macro gain is concentrated in the sparse movements rather than uniform, and it regresses on two classes (Vorticism $-$13 at $n{=}8$, and the saturated Renaissance $-$3), which we attribute to small support pools and visual confusability. Scaling the support pool for under-represented movements is the most direct lever for closing the remaining gap.

\begin{table}[tb]
\caption{Per-movement top-1 accuracy on the 236-query expanded set (GPT-4.1), vanilla vs.\ \method. Gains concentrate on the sparse, less canonical movements; Abstract Expressionism is the one genuinely hard class.}
\centering
\small
\setlength{\tabcolsep}{4pt}
\begin{tabular}{lccc}
\toprule
Movement & $n$ & vanilla & \method \\
\midrule
Post-Impressionism     & 42 & 0.79 & 0.83 \\
Baroque                & 39 & 0.72 & 0.82 \\
Impressionism          & 35 & 0.74 & 0.80 \\
Renaissance            & 31 & 1.00 & 0.97 \\
Abstract Expressionism & 25 & 0.04 & 0.04 \\
Romanticism            & 12 & 0.58 & 0.83 \\
Fauvism                & 11 & 0.45 & 0.73 \\
Pointillism            &  9 & 0.33 & 0.78 \\
Futurism               &  8 & 0.38 & 0.62 \\
Suprematism            &  8 & 0.38 & 0.62 \\
Surrealism             &  8 & 0.88 & 0.88 \\
Vorticism              &  8 & 0.38 & 0.25 \\
\bottomrule
\end{tabular}
\label{tab:permovement}
\end{table}

\subsection{Shot count and embedder choice}
\label{sec:exp-extra-ablations}

Two further ablations confirm robustness, with full tables in Appendix~\ref{sec:supp-ablations}. A shot-count sweep ($K\in\{0,1,3,5,10\}$) for GPT-4.1 with \method on \bench shows macro accuracy jumping 3.1 points from $K{=}0$ to $K{=}1$, then staying essentially flat through $K{=}10$ (range 0.751--0.762). This agrees with the random-versus-cosine ablation (\S\ref{sec:exp-ablation}): the load-bearing factor is having structured retrieval at all rather than the shot count. We use $K{=}5$ as the default, the best macro by 0.2--1.1 points while well below the per-query token-cost ceiling. Swapping the default Qwen3-VL-Embedding-8B retriever for CLIP-ViT-L/14~\cite{radford2021clip} or the much smaller SigLIP-base~\cite{zhai2023siglip} (90M) holds the gain (2.6 to 4.5 points macro; SigLIP-base in fact reaches 0.766, marginally above the 8B default's 0.762), so on GPT-4.1 \method is robust to embedder choice within the multimodal family and a cheap 90M embedder suffices. We keep the 8B default for its label-conditioned encoding $\phi(x_s,y_s)$ and cross-task robustness, and leave the open Qwen3-VL-30B's embedder sensitivity to future work.

\subsection{Cross-task generality: Bongard-OpenWorld}
\label{sec:exp-bongard}

To test whether retrieval-as-thinking transfers beyond definition-grounded classification, we apply the same \method scaffold to Bongard-OpenWorld~\cite{wu2024bongardow}, a free-form concept-induction benchmark in which each problem gives 6 positive and 6 negative images of an unstated concept and the model decides whether a held-out query satisfies it, with no symbolic labels. Only the turns layer is active: the twelve examples become alternating turns (the assistant says \texttt{positive} or \texttt{negative}) and the query is the final user turn. Vanilla GPT-4.1 sits at 0.466, at the binary-chance line, since without exemplars the model has no signal on the unstated concept, and the scaffold lifts it to 0.727 (+26.1 points; Tab.~\ref{tab:bongard}), reported strictly as a controlled same-backbone delta. The contrast isolates the turns mechanism, since the backbone, query, and prompt are identical and only the twelve example turns differ. We report this as a diagnostic transfer on the 44 of 200 problems whose 14 images all decoded from the published URLs, a fetch filter independent of concept difficulty (Appendix~\ref{sec:supp-bongard}), so the published GPT-4V score of 0.640 (a different model on the full 200-problem set) is a reference point rather than a head-to-head. That the same scaffold lifts a task with no taxonomy positions \method as a retrieval-based baseline for the Thinking-with-Images programme.

\begin{table}[!tb]
\caption{Binary accuracy on Bongard-OpenWorld~\cite{wu2024bongardow} (44 problems, 88 queries). Published GPT-4V from the original benchmark for reference. \method, with zero task-specific tuning, lifts GPT-4.1 by $+26.1$~pts.}
\centering
\small
\setlength{\tabcolsep}{6pt}
\begin{tabular}{lcc}
\toprule
Method & Binary Acc. & $\Delta$ vs.\ vanilla \\
\midrule
\multicolumn{3}{l}{\emph{Published~\cite{wu2024bongardow}}} \\
GPT-4V (published)                  & 0.640 & --      \\
Human performance                   & 0.91  & --      \\
\midrule
\multicolumn{3}{l}{\emph{Ours}} \\
GPT-4.1 (vanilla, no examples)      & 0.466 & --      \\
\textbf{GPT-4.1 + \method}          & \textbf{0.727} & \textbf{$+26.1$} \\
\bottomrule
\end{tabular}
\label{tab:bongard}
\end{table}

\section{Conclusion}
\label{sec:conclusion}

We presented \method, a train-free framework for Thinking with Images that occupies the retrieval-based position in its design space, built from three composable layers (structured definitions, per-query multimodal retrieval, and alternating-turns injection) with no fine-tuning. It degrades gracefully to whichever layers a task admits: it ties the 72B LLaVA-OneVision SOTA on VL-ICL at about 1/2.4 the parameters using retrieval and turns, yields a 4--6 point macro gain across three backbones on \bench using the full stack, and adds 26.1 points on Bongard-OpenWorld using turns alone. A same-task build-up shows that retrieval and turns are the universal lever while structured definitions are need-adaptive, and ablations attribute 83\% of the \bench retrieval gain to which exemplars are retrieved.

\paragraph{Limitations.}
\bench's core split is compact (153 images) and positioned as a diagnostic; we report bootstrap intervals and expand the curator-derived attributes to 236 queries, where all three backbones stay significant with tighter intervals (Appendix~\ref{sec:supp-stats}). Composition and Mood stay at the 96 diagnostic queries with only moderate agreement ($\kappa_{\max}{=}0.43/0.54$), and a Claude Opus~4.8 re-derivation leaves the curator-derived majority gold unchanged, though a residual influence on the two subjective attributes remains possible. Our claim of narrowing the open-versus-closed gap is benchmark-specific: the open Qwen3-VL-30B ties the 72B SOTA on VL-ICL but still trails GPT-4.1 by 7.3 points on \bench, so the gap closes only in the synthetic-name regime. Finally, \method targets the small-shot regime and inherits the VLM's long-context degradation at large $K$, and we do not test fine-tuning or multi-step reasoning-chain tasks where generation-based methods may retain an edge.

\bibliography{main}

\appendix
\section*{Appendix}
\noindent The following material supplements the main paper. It was the separate
supplementary document in the AAAI submission and is included here so the arXiv
version is self-contained; all tables, figures, and sections are referenced from
the main text at the point where they are summarized.
\section{\bench dataset statistics}
\label{sec:supp-stats}

\bench comprises 153 public-domain images (96 query / 57 support) drawn from three CC0/PD sources: Metropolitan Museum of Art Open Access (55), National Gallery of Art Open Access (45), and Wikimedia Commons PD-in-source-country (53). Five attributes are annotated per image: \textbf{Movement} (12 classes), \textbf{Genre} (6), \textbf{Medium} (5 defined; 4 attested in the released split), \textbf{Composition} (4), and \textbf{Mood} (5). Movement, Genre, and Medium are curator-derived; Composition and Mood are 3-VLM ensemble labels passed through the rubric-grounded verifier (\S\ref{sec:bench-verify}). Table~\ref{tab:supp-splits} gives the per-movement query/support split: a uniform $8/5$ per movement, except Surrealism ($8/2$) where the strict tier-2 license-and-date policy admits only 10 qualifying Wikimedia images.

\begin{table}[h]
\caption{Per-movement query/support split in \bench. Total: 96 query, 57 support.}
\centering
\small
\setlength{\tabcolsep}{4pt}
\begin{tabular}{lcc}
\toprule
Movement & Query & Support \\
\midrule
Renaissance, Baroque, Romanticism, & & \\
Impressionism, Post-Impressionism, & 8 & 5 \\
Pointillism, Fauvism, Futurism, & (each) & (each) \\
Suprematism, Abstract Expr., Vorticism & & \\
\midrule
Surrealism (metaphysical) & 8 & 2 \\
\bottomrule
\end{tabular}
\label{tab:supp-splits}
\end{table}
\FloatBarrier

\paragraph{Expanded curator-derived split.}
To test small-sample robustness (\S\ref{sec:findings}), we enlarge the three curator-derived attributes (Movement, Genre, Medium) from 96 to \textbf{236 query images} by adding 140 additional license-clean works: 71 from the Metropolitan Museum (CC0) and 69 from the National Gallery of Art (CC0), bringing \bench to 293 images total (Met 126 / NGA 114 / Wikimedia 53). Labels are assigned with the same provenance as the core split: Movement through a verified artist$\to$movement lookup (100\% recovery on the original Met/NGA reference rows), Medium through a curator-classification rule (98\%), and Genre through Getty-AAT tag rules (93\%), with the few residual cases resolved by textbook-certain attributions. The added images carry no Composition/Mood gold, so those two attributes remain at the original 96 diagnostic queries; the support pool (57, 5-attribute) is left unchanged so that in-context exemplar demonstrations stay consistent. Table~\ref{tab:supp-expanded} gives the expanded per-movement counts. The expansion is deliberately not re-balanced: it concentrates on canonical movements (Baroque, Post-Impressionism, Impressionism, Renaissance), where public-domain holdings are dense, while the avant-garde movements (Futurism, Suprematism, Surrealism, Vorticism) stay at the original 8 queries because licence-clean holdings are scarce. The gain remains significant for every backbone despite this skew (\S\ref{sec:findings}).

\begin{table}[h]
\caption{Expanded vs.\ core per-movement query counts in \bench (curator-derived attributes). The support pool is unchanged at 57.}
\centering
\small
\setlength{\tabcolsep}{4pt}
\begin{tabular}{lcc}
\toprule
Movement & Core & Expanded \\
\midrule
Renaissance            & 8 & 31 \\
Baroque                & 8 & 39 \\
Romanticism            & 8 & 12 \\
Impressionism          & 8 & 35 \\
Post-Impressionism     & 8 & 42 \\
Pointillism            & 8 & 9  \\
Fauvism                & 8 & 11 \\
Abstract Expressionism & 8 & 25 \\
Futurism               & 8 & 8  \\
Suprematism            & 8 & 8  \\
Surrealism             & 8 & 8  \\
Vorticism              & 8 & 8  \\
\midrule
Total                  & 96 & 236 \\
\bottomrule
\end{tabular}
\label{tab:supp-expanded}
\end{table}
\FloatBarrier

\section{Component build-up (full table)}
\label{sec:supp-buildup}
This table backs the build-up analysis in \S\ref{sec:exp-buildup} (main text). Layers are added top-to-bottom on \bench; the retrieval+turns layer is the dominant, backbone-universal lever, while structured definitions are need-adaptive.

\begin{table}[h]
\caption{Same-task component build-up on \bench (macro accuracy, reparsed gold). The retrieval+turns layer is the dominant, universal lever ($+4.1$/$+5.4$); structured definitions are \emph{need-adaptive} --- they help the open Qwen3-VL-30B ($+2.5$), which lacks internalized art-historical knowledge, but are mildly redundant for GPT-4.1 ($-1.1$), which already knows the taxonomy.}
\centering
\small
\setlength{\tabcolsep}{4pt}
\begin{tabular}{lcc}
\toprule
Layers active & GPT-4.1 & Qwen3-VL-30B \\
\midrule
Bare (label space only)              & 0.732 & 0.610 \\
\;+ structured definitions           & 0.721 \footnotesize{($-1.1$)} & 0.635 \footnotesize{($+2.5$)} \\
\;+ retrieval + turns (full \method) & \textbf{0.762} \footnotesize{($+4.1$)} & \textbf{0.689} \footnotesize{($+5.4$)} \\
\bottomrule
\end{tabular}
\label{tab:buildup}
\end{table}
\FloatBarrier

\section{Per-movement and provenance details}
\label{sec:supp-permovement}

The per-movement vanilla-versus-\method accuracy on the 236-query set (Table~\ref{tab:permovement}) and the \bench accuracy split by label provenance (Table~\ref{tab:provenance}) are both reported in the main text. The provenance split follows best practice for benchmarks that mix human-curated and model-derived labels: the curator-derived attributes (Movement, Genre, Medium, from Met/NGA/WikiArt metadata) use the 236-query expanded set, while the model-derived attributes (Composition, Mood, from the 3-model ensemble) stay at the 96-query diagnostic set and carry lower agreement, so \method's gains are reported and tested only on the curator-derived attributes.

\section{Per-attribute inter-annotator agreement}
\label{sec:supp-kappa}

\begin{figure}[h]
    \centering
    \includegraphics[width=0.95\linewidth]{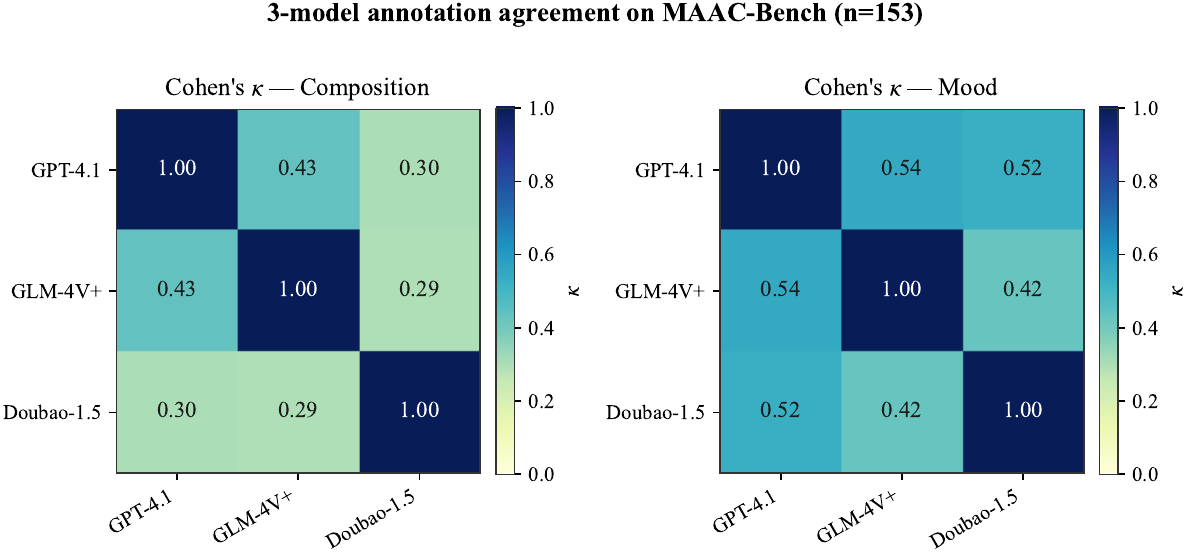}
    \caption{Pairwise Cohen's $\kappa$ between the three annotator VLMs (GPT-4.1, GLM-4V-Plus, Doubao-1.5-vision-pro-32k) on \bench for Composition (left) and Mood (right). GPT-4.1 $\leftrightarrow$ GLM-4V-Plus has the highest agreement on both attributes.}
    \label{fig:supp-kappa}
\end{figure}
\FloatBarrier

\section{Per-movement gold coverage}
\label{sec:supp-splits}

\begin{figure}[h]
    \centering
    \includegraphics[width=0.95\linewidth]{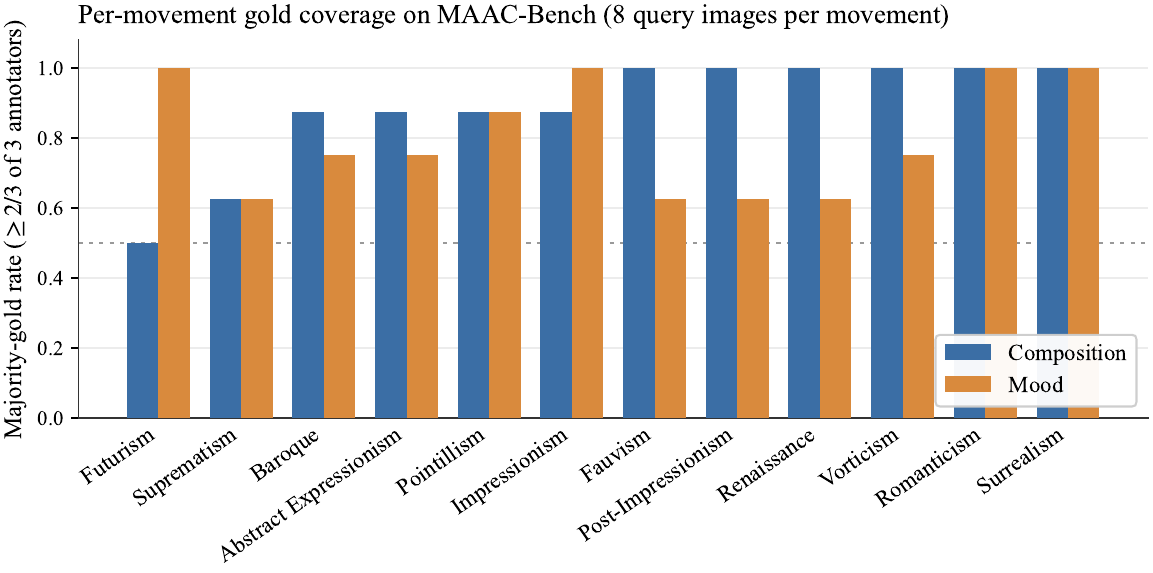}
    \caption{Per-movement majority-gold coverage (fraction of the 8 query images per movement where ${\geq}2/3$ models agreed on a label). Composition coverage degrades on Futurism (4/8) and Suprematism (5/8) because the 4-class composition rubric was designed for figurative work and does not transfer cleanly to abstract movements.}
    \label{fig:supp-goldcov}
\end{figure}
\FloatBarrier

\section{Per-attribute full headline figure}
\label{sec:supp-headline}

\begin{figure}[h]
    \centering
    \includegraphics[width=0.95\linewidth]{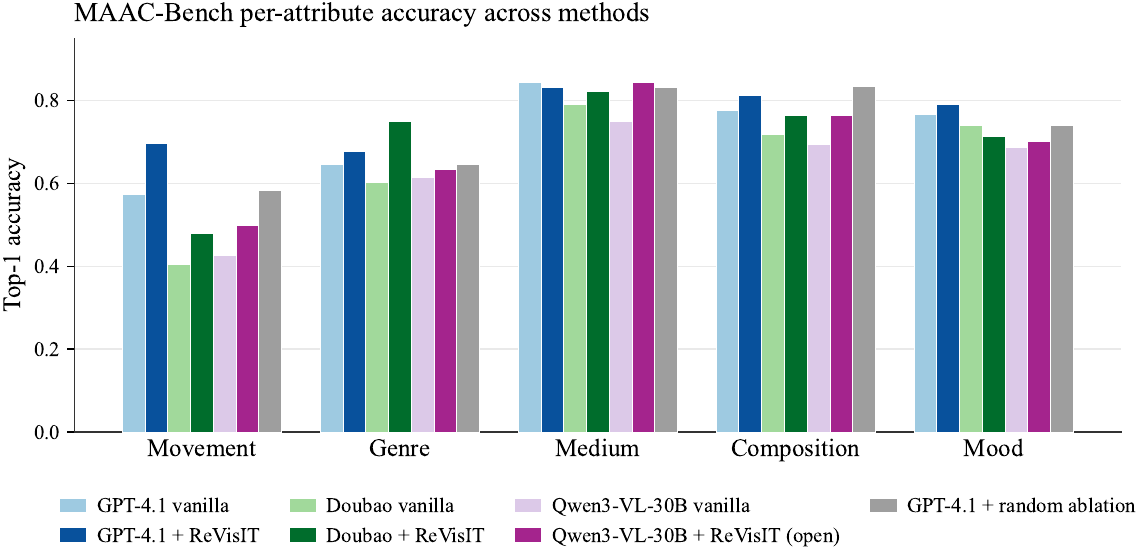}
    \caption{Per-attribute top-1 accuracy on \bench, all 7 methods. Dark vs.\ light bars of the same color show the marginal contribution of \method's retrieval + in-context exemplar layer on each underlying VLM.}
    \label{fig:supp-headline}
\end{figure}
\FloatBarrier

\section{Claim-to-evidence map}
\label{sec:supp-claims}
\begin{table}[h]
\caption{Each headline claim mapped to its supporting evidence.}
\centering
\small
\setlength{\tabcolsep}{3pt}
\begin{tabular}{p{0.45\linewidth}p{0.48\linewidth}}
\toprule
Claim & Evidence \\
\midrule
Retrieval is the load-bearing layer & embedding-only probes underperform the VLM scaffold (kNN-vote $0.562$ vs.\ \method $0.799$, Tab.~\ref{tab:probe}); cosine-vs-random isolates $83\%$ on \bench (\S\ref{sec:exp-ablation}); turns $\approx$ flat (\S\ref{sec:exp-ablation}) \\
Gain is backbone-portable & 3 architecturally diverse backbones; paired bootstrap CIs exclude 0 on curator attributes (\S\ref{sec:findings}) \\
Gain is leakage-robust for the open backbone & same-artist-free subset: $+5.5$~pt for Qwen (preserved); GPT-4.1 shrinks to $+1.6$~pt (\S\ref{sec:findings}, F4); zero exact-ID overlap \\
Open 30B matches 72B (param-efficiency) & VL-ICL $0.985$ vs.\ $0.987$ at $\sim$1/2.4 params (Tab.~\ref{tab:vlicl}) \\
Definitions are need-adaptive & $+2.5$ for open Qwen, $-1.1$ for GPT-4.1 (Tab.~\ref{tab:buildup}) \\
\bottomrule
\end{tabular}
\label{tab:claims}
\end{table}
\FloatBarrier

\section{Inference pipeline pseudocode}
\label{sec:supp-algo}

\begin{algorithm}[h]
\caption{The \method inference pipeline (\S\ref{sec:method})}
\label{alg:revisit}
\begin{algorithmic}[1]
\REQUIRE Query $x_q$; taxonomy $\mathcal{T}$; support pool $\mathcal{S}$; embedder $\phi$; VLM $\pi$; shot count $K$
\STATE \textbf{[Offline]} $\forall (x_s, y_s)\in\mathcal{S}: e_s \gets \phi(x_s, y_s)$
\STATE $\mathcal{E}_K \gets \mathrm{topK}_{s\in\mathcal{S}}\,\cos(\phi(x_q),\, e_s)$ \hfill\COMMENT{retrieval}
\STATE $M \gets [\mathrm{Compile}(\mathcal{T})]$ \hfill\COMMENT{three-block sys prompt}
\STATE \textbf{for} $(x_e, y_e)\in\mathcal{E}_K$: append $\langle$user:$x_e\rangle,\langle$asst:$y_e\rangle$ to $M$
\STATE Append $\langle$user:$x_q\rangle$ to $M$
\STATE $J \gets$ greedy-decode $\pi(M)$, parse as JSON
\STATE $\forall$ predicted $\ell$: $\mathrm{conf}(\ell) \gets \mathrm{softmax}(\mathrm{logits}_t)[\mathrm{argmax}]$ where $t$ is the first step containing $\ell$
\STATE Filter $J$ to per-category whitelist; truncate to $\mathrm{limit}(c_i)$ labels.
\RETURN $J,\ \{\mathrm{conf}(\ell)\}$
\end{algorithmic}
\end{algorithm}
\FloatBarrier

\section{Bongard-OpenWorld protocol details}
\label{sec:supp-bongard}

Each Bongard-OpenWorld problem provides 6 positive and 6 negative example images of an \emph{unstated} concept; the model must decide whether a held-out query image satisfies it. We download the test split via the published URLs (200 problems) and retain the 44 problems whose 12 example images and 2 queries \emph{all decoded successfully} (88 binary predictions). This $\sim$22\% retention reflects \emph{download/decode failures of the published image URLs}, not difficulty-based selection: a problem is dropped iff one of its 14 images failed to fetch --- a process independent of the underlying concept. We therefore report this as a \emph{diagnostic transfer} result and treat the published GPT-4V number as a reference point rather than a controlled head-to-head. \method's adaptation is automatic: the alternating-turns template ingests the 6+6 examples (assistant says \texttt{positive}/\texttt{negative}); the query becomes the final user turn. No taxonomy, definitions, or retrieval embedder are invoked --- the only inductive bias is the alternating-turns scaffold itself.

\section{Shot-count and embedder ablations (full tables)}
\label{sec:supp-ablations}

These tables back the compact summary in \S\ref{sec:exp-extra-ablations}.

\paragraph{K-shot sweep.}
Table~\ref{tab:kshot} sweeps $K\in\{0,1,3,5,10\}$ for GPT-4.1 + \method on \bench. Adding even a single retrieved exemplar already captures most of the gain; the macro accuracy is essentially flat from $K{=}1$ onward.

\begin{table}[h]
\caption{K-shot sweep on \bench (GPT-4.1 + \method). Adding even a single retrieved exemplar already captures most of the gain. \emph{Macro} is averaged over all five attributes; the Composition column is omitted here for width and is reported in Tab.~\ref{tab:headline}, so the four displayed columns do not by themselves average to Macro.}
\centering
\small
\setlength{\tabcolsep}{4pt}
\begin{tabular}{cccccc}
\toprule
$K$ & Macro & Movement & Genre & Medium & Mood \\
\midrule
0 (vanilla) & 0.721 & 0.573 & 0.646 & \textbf{0.844} & 0.766 \\
1           & 0.752 & 0.615 & 0.677 & 0.854 & 0.779 \\
3           & 0.751 & 0.615 & 0.688 & 0.865 & 0.753 \\
\textbf{5}  & \textbf{0.762} & \textbf{0.698} & 0.677 & 0.833 & \textbf{0.792} \\
10          & 0.760 & 0.646 & 0.688 & \textbf{0.865} & \textbf{0.792} \\
\bottomrule
\end{tabular}
\label{tab:kshot}
\end{table}
\FloatBarrier

\paragraph{Retrieval embedder choice.}
Table~\ref{tab:embedder} swaps the default Qwen3-VL-Embedding-8B retriever for CLIP-ViT-L/14~\cite{radford2021clip} and SigLIP-base~\cite{zhai2023siglip}, holding the downstream VLM (GPT-4.1) and prompt format constant. All three multimodal embedders yield a positive macro gain ($+2.6$ to $+4.5$~pts), and the differences across embedders are smaller than the gain itself --- notably the 90M SigLIP-base matches the 8B default. \method is robust to embedder choice within the multimodal family and does not require a large dedicated VLM embedder.

\begin{table}[h]
\caption{Retrieval embedder ablation on \bench (GPT-4.1 + \method, $K{=}5$). All three embedders yield a positive gain; the framework is robust to embedder choice within the multimodal family.}
\centering
\small
\setlength{\tabcolsep}{5pt}
\begin{tabular}{lcc}
\toprule
Retrieval embedder & Macro & $\Delta$ vs.\ vanilla \\
\midrule
None (vanilla, $K{=}0$)              & 0.721 & --        \\
CLIP-ViT-L/14 (300M)                  & 0.747 & $+2.6$    \\
\textbf{SigLIP-base (90M)}            & \textbf{0.766} & \textbf{$+4.5$}  \\
Qwen3-VL-Embedding-8B (default)       & 0.762 & $+4.1$    \\
\bottomrule
\end{tabular}
\label{tab:embedder}
\end{table}
\FloatBarrier

\section{Reproducibility checklist}
\label{sec:supp-repro}
\textbf{Models.} GPT-4.1 and Doubao-1.5-vision-pro-32k via the friday OpenAI-compatible gateway (accessed 2026-06); Qwen3-VL-30B-A3B-Instruct (open-weights MoE, \texttt{Qwen3VLMoeForConditionalGeneration}) on a single A100~80GB. Composition/Mood independent re-annotation uses Claude Opus~4.8.
\textbf{Retriever.} Qwen3-VL-Embedding-8B; cosine similarity; top-$K{=}5$.
\textbf{Decoding.} Greedy (temperature $0$); \texttt{max\_new\_tokens}$=512$; identical structured three-block prompt for vanilla and \method (vanilla drops only the exemplars).
\textbf{Image preprocessing.} Max-edge $1200$px, JPEG.
\textbf{Software and hardware.} Local Qwen3-VL-30B-A3B and Qwen3-VL-Embedding-8B inference runs through the HuggingFace Transformers \texttt{generate} path (SDPA attention) on a single NVIDIA A100-SXM4-80GB, under Python~3.10, PyTorch~2.8.0 (CUDA~12.8, cuDNN~9.10), Transformers~4.57.1, and NumPy~1.26.4, on CentOS~7 (Linux~4.18). GPT-4.1, Doubao-1.5, and GLM-4V-Plus are accessed as hosted APIs.
\textbf{Determinism and seeds.} The \method pipeline is deterministic: greedy decoding and cosine top-$K$ retrieval introduce no randomness. The only stochastic component is the random-retrieval ablation, whose exemplars are drawn with NumPy \texttt{default\_rng} seeded per query as $42+i$ for query index $i$, making that ablation exactly reproducible.
\textbf{Data.} 96-query / 57-support core split plus a 236-query curator-derived expansion; released with per-image object IDs, source URLs, and license tags. Composition/Mood gold is a 3-model ensemble majority with a rubric-grounded verifier.
\textbf{Statistics.} Paired bootstrap, 10{,}000 resamples; 95\% percentile intervals.

\end{document}